\documentclass[10pt,letterpaper]{article}
\usepackage{graphicx}
\usepackage{cogsci}

\cogscifinalcopy 

\usepackage[
  style=apa,
  natbib=true,
  annotation=false,
]{biblatex}
\usepackage{subcaption}
\usepackage{booktabs}
\usepackage{array}
\usepackage{multirow}
\usepackage{multicol}
\usepackage{rotating}
\usepackage{placeins}
\usepackage{dblfloatfix}
\usepackage{url}

\newcommand{\keywords}[1]{\textbf{Keywords:} #1}

\usepackage{float} 

\title{Objects Before Words: Object-First Inductive Biases for Grounding Language in Child-View Video}

\author{
  {\large\bfseries
    Sathira Silva (sathira.silva@mbzuai.ac.ae)$^1$, 
    Abrham Kahsay Gebreselasie$^1$, 
    Muhammad Umer Sheikh$^1$, 
    Kartik Kuckreja$^1$, 
    Daniel Harari$^2$, \&
    Muhammad Haris Khan$^2$
  } \\
  {\normalsize\normalfont
    $^1$Mohamed bin Zayed University of Artificial Intelligence, Abu Dhabi, UAE \\
    $^2$Weizmann Institute of Science, Rehovot, Israel
  }
}

\bibliography{main}

\begin{document}

\maketitle

\begin{abstract}
Learning grounded word meaning from natural experience requires resolving two ambiguities in infant-view recordings: \emph{when} the named referent appears and \emph{where} it is in a cluttered frame. In SAYCam-style data, caregiver speech is sparse and weakly synchronized with egocentric video, so single-frame contrastive pairing yields noisy positives in which the intended object is absent or entangled with distractors. We propose \textsc{BabyMind}, an object-first bias for child-view contrastive learning under sparse, noisy supervision. \textsc{BabyMind} extracts candidate object embeddings using an offline mask-based region interface, links candidates across a short utterance-centered window into lightweight object files via tracking, and aligns utterances to bags of object files with a prototype-space multiple-instance contrastive objective. Track-coherence and global-object agreement regularizers stabilize learning and transfer object-file structure into the global frame embedding used at evaluation. On SAYCam-S, \textsc{BabyMind} improves Labeled-S 15 forced-choice accuracy by +2.6 points over CVCL and yields consistent gains on in-vocabulary out-of-distribution benchmarks. We release our codes at \url{https://github.com/sathiiii/BabyMind}.

\keywords{grounded language learning; child-view video; egocentric vision; contrastive learning; multiple-instance learning; object files; prototype memory; SAYCam}
\end{abstract}

\section{Introduction}

\begin{figure}[t]
    \centering
    \begin{subfigure}[t]{0.49\linewidth}
        \centering
        \includegraphics[width=\linewidth]{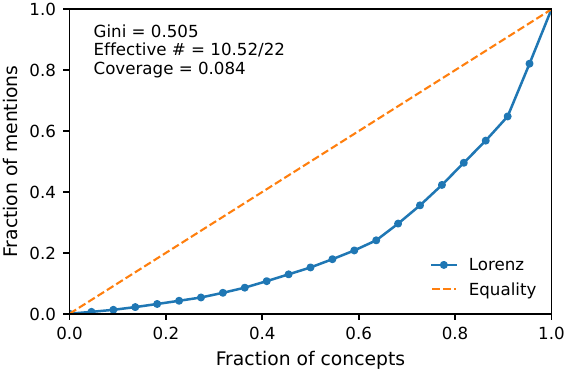}
        \caption{}
        \label{fig:sparsity_lorenz}
    \end{subfigure}
    \hfill
    \begin{subfigure}[t]{0.49\linewidth}
        \centering
        \includegraphics[width=\linewidth]{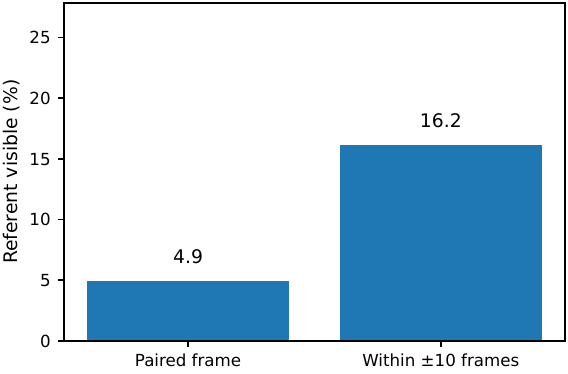}
        \caption{}
        \label{fig:sparsity_visibility}
    \end{subfigure}
    \vfill
    \begin{subfigure}[t]{\linewidth}
        \centering
        \includegraphics[width=\linewidth]{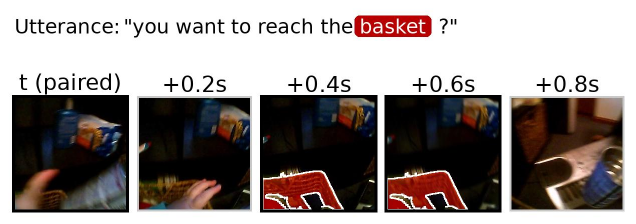}
        \caption{}
        \label{fig:temporal_misalignment}
    \end{subfigure}

    \caption{\textbf{SAYCam-S sparsity and misalignment.}
    (a) Long-tailed concept mentions.
    (b) Referents are often absent in the paired frame but present within a short window.
    (c) Example where the referent appears shortly after the paired time step.}
    \label{fig:concept_sparsity}
\end{figure}

A central goal of grounded language learning is to explain how a learner can acquire word meanings from perceptual experience paired with sparse, weakly structured linguistic input \citep{Harnad1990SymbolGrounding}. This problem looks fundamentally different in early child learning than in curated image-caption corpora: the visual stream is egocentric, cluttered, partially occluded, and constantly moving, while caregiver speech is intermittent and only loosely synchronized with what is currently in view.
In SAYCam-style data \citep{Sullivan2021SAYCam}, this creates two recurring ambiguities: \emph{when} the named referent appears and \emph{where} it is in a crowded scene.
These are not edge cases: concept mentions are long-tailed and imbalanced (Figure~\ref{fig:concept_sparsity}\subref{fig:sparsity_lorenz}), and a referent can be absent at the paired time step but present nearby in time (Figure~\ref{fig:concept_sparsity}\subref{fig:sparsity_visibility}), as illustrated qualitatively in Figure~\ref{fig:concept_sparsity}\subref{fig:temporal_misalignment}.
These properties motivate an ``objects before words'' inductive bias.
Classic accounts argue that robust recognition depends on intermediate perceptual structure that supports later interpretation \citep{Marr1982Vision,Mandler1992ConceptualPrimitives}.
Empirically, infants track object continuity under motion and occlusion and often generalize early nouns by shape \citep{KellmanSpelke1983Occlusion,Baillargeon1987ObjectPermanence,Landau1988Importance,Smith2003ShapeBiasReview}.
By contrast, standard visual representations can over-rely on texture and background cues \citep{Geirhos2018ImagenetBias}, which is especially problematic in egocentric scenes where context is pervasive and named referents can occupy only a small region. Child-View Contrastive Learning (CVCL) \citep{Vong2024GroundedLanguage} takes an important step toward this setting by learning CLIP-style cross-modal embeddings \citep{Radford2021CLIP} from utterances paired with child-view frames.
However, CVCL trains on a single sampled frame per utterance, so positives can be noisy under temporal jitter, occlusion, and clutter.
Whole-frame embeddings can also explain an utterance using background context or salient distractors, which can be especially harmful for rare concepts (Figure~\ref{fig:concept_sparsity}\subref{fig:sparsity_lorenz}).

Motivated by this perspective, we introduce \textsc{BabyMind}, an object-first inductive bias that augments CVCL with an auxiliary object-file pathway.
\textsc{BabyMind} keeps the original CVCL global contrastive objective on the same single anchor frame per utterance and adds a short window of nearby frames used only for object-centric learning.
From each frame, we extract \emph{instance-level} candidates using an offline automatic mask generation (AMG) strategy, with a patch fallback to handle missing or degenerate masks.
We then link candidates across the window into short \emph{object files} using lightweight tracking and align the utterance to the resulting \emph{bag of tracked object files} with a multiple-instance contrastive objective \citep{Maron1998MIL}.
Crucially, this is not just relaxing a one-to-one pairing into a one-of-many relation: the latent alignment is constrained to tracked, instance-defined candidates and further stabilized by a small prototype memory plus auxiliary signals.
Specifically, we compute MIL in prototype space (a learned codebook that encourages reusable appearance structure under long-tailed, noisy supervision), regularize tracks with a coherence loss across frames, and add a global-object agreement loss that transfers the selected object-file signal into the global frame embedding used at evaluation. Our contributions can be summarized as:
\begin{itemize}
    \item \textsc{BabyMind}, an object-first extension of CVCL that addresses temporal and spatial ambiguity by aligning speech to tracked instance candidates within a short window, while keeping the original CVCL global objective and evaluation interface.
    \item Offline Automatic Mask Generation (AMG) for instance-level region masks and short-window object files via lightweight tracking for egocentric video.
    \item A prototype-space multiple-instance contrastive objective, with track coherence and global-object agreement to stabilize object selection and inject object-centric structure into the global embedding.
    \item Improved SAYCam-S grounding on Labeled-S 15 and consistent (if modest) gains on IV out-of-distribution evaluations under the CVCL protocol.
\end{itemize}

\section{Related Work}

\paragraph{Grounded language learning from child-view video.}
Long-form egocentric corpora such as SAYCam provide a naturalistic testbed for grounded learning under clutter, occlusion, and weak temporal coupling between caregiver speech and the child’s visual stream \citep{Sullivan2021SAYCam}.
CVCL demonstrates that CLIP-style vision-language alignment can be learned in this regime using utterance-frame pairing and contrastive objectives \citep{Vong2024GroundedLanguage,Radford2021CLIP}, complementing evidence that high-level visual representations can emerge from child-like inputs even with limited built-in structure \citep{OrhanGL20,OrhanLake2024ChildPerspective}.
A central challenge in this setting is that supervision is intrinsically ambiguous: the relevant object may fall outside the sampled frame or be visually confounded by background context.
Complementary benchmark evidence \citep{Chen2026BabyVision} suggests that perceptual competence remains a bottleneck for current multimodal systems even on tasks that do not primarily depend on language, motivating stronger perceptual organization for grounding.

\paragraph{Object-centric structure and region interfaces in vision-language.}
A large body of vision-language work injects object-level structure via region proposals or detector features, which has become a standard interface for captioning/VQA and later for region-aware pretraining \citep{Anderson2018BottomUp,Tan2019LXMERT,Li2020Oscar,Zhang2021VinVL}.
Related ideas appear in self-supervised learning, where region/mask structure is used to reduce shortcut reliance and to define localized learning targets \citep{Henaff2021DetCon}.
In parallel, object-centric representation learning has explored architectural constraints that decompose scenes into entities without labels \citep{Burgess2019MONet,Locatello2020ObjectCentric}.
Recent foundation segmentation models make it feasible to obtain instance masks as a general-purpose perceptual prior across domains \citep{Kirillov2023SegmentAnything}, aligning with results showing that perceptual inductive biases can materially shape what contrastive learning captures \citep{PerceptualBiasCL}.

\paragraph{Ambiguous instance selection, prototype memories, and temporal persistence.}
When supervision applies to a set of candidates rather than a single labeled instance, multiple-instance learning (MIL) provides a principled framework for latent instance selection \citep{Maron1998MIL}.
Prototype- and clustering-based mechanisms are also widely used to stabilize learning and encourage reusable structure in self-supervision, including memory-bank approaches and online assignment/clustering methods \citep{Wu2018InstanceDiscrimination,Caron2018DeepCluster,Caron2020SwAV,Caron2021DINO,Oord2017VQVAE}.
Finally, learning from temporal continuity is a longstanding theme in unsupervised learning from video \citep{Wiskott2002SFA,Sermanet2017TCN}, and cognitive accounts formalize persistence via token-like “object files” maintained across time \citep{Kahneman1992ObjectFiles}.
These threads motivate combining set-based alignment with reusable prototype structure and short-range temporal persistence when learning grounded meaning from egocentric video.

\section{Methodology}
\label{sec:methodology}

\paragraph{Overview.}
We learn aligned representations of child-view video and caregiver speech in the Child-View Contrastive Learning (CVCL) setting~\citep{Vong2024GroundedLanguage} on SAYCam~\citep{Sullivan2021SAYCam}.
In egocentric video, an utterance may refer to an object whose time of appearance and pixels are unknown, so single-frame contrastive pairing is sensitive to temporal mismatch and background clutter.
\textsc{BabyMind} resolves this ambiguity by introducing an object-file pathway inspired by token-level persistence~\citep{Kahneman1992ObjectFiles}:
(i) a fixed region interface from offline automatic mask generation (AMG)~\citep{Kirillov2023SegmentAnything},
(ii) short-window object files formed by greedy cross-frame linking,
and (iii) a prototype-space multiple-instance contrastive objective~\citep{Maron1998MIL}.
Two regularizers shape this pathway: track coherence in prototype space (temporal stability~\citep{Wiskott2002SFA}) and global-object agreement that transfers object-file structure into the global embedding used at evaluation.
\begin{figure*}[t]
    \centering
    \includegraphics[trim={5cm 4cm 5cm 5cm},width=0.85\linewidth]{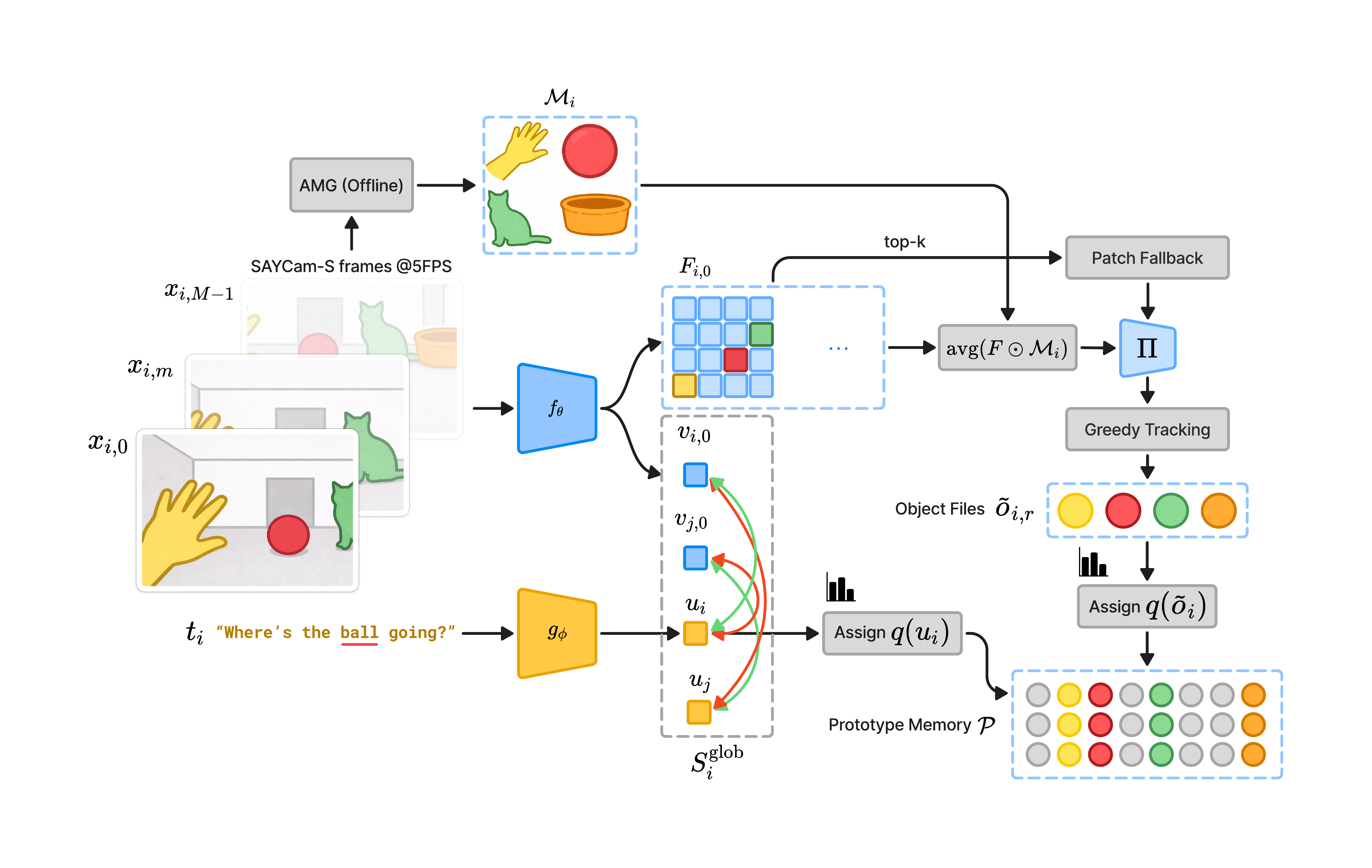}
    \caption{\textbf{Method overview.}
    Each example contains an utterance $t_i$ and a window of $M$ utterance-aligned frames $\{x_{i,m}\}_{m=0}^{M-1}$, with $m=0$ the anchor used by the global CVCL objective.
    We extract mask-defined region embeddings from feature maps, merge them into object files across the window, align text to the resulting bag via prototype-space MIL, regularize tracks via prototype consistency across frames, and align the anchor global embedding to a prototype reconstruction of the selected object file.}
    \label{fig:method_overview}
\end{figure*}
\subsection{Problem setup}
A minibatch contains $B$ examples indexed by $i\in\{1,\dots,B\}$.
Example $i$ contains an utterance transcript $t_i$ and a temporal window of $M$ frames $\{x_{i,m}\}_{m=0}^{M-1}$.
The anchor frame $x_{i,0}$ is sampled uniformly at random from the utterance-aligned frame set as in CVCL and is used by
$\mathcal{L}_{\mathrm{glob}}$.
The remaining $M{-}1$ frames are additional samples from the same utterance-aligned set and are used only for object-file learning.
A text encoder $g_{\phi}$ and a vision encoder $f_{\theta}$ produce normalized embeddings and a spatial feature map:
\begin{align}
u_i &= \frac{g_{\phi}(t_i)}{\|g_{\phi}(t_i)\|_2} \in \mathbb{S}^{d-1},\\
v_{i,m} &= \frac{f_{\theta}^{\mathrm{glob}}(x_{i,m})}{\|f_{\theta}^{\mathrm{glob}}(x_{i,m})\|_2} \in \mathbb{S}^{d-1},\\
F_{i,m} &= f_{\theta}^{\mathrm{map}}(x_{i,m}) \in \mathbb{R}^{C\times H\times W}.
\end{align}
We denote $v_i=v_{i,0}$ and use cosine similarity $s(a,b)=a^\top b$ for normalized vectors.

\subsection{Global CVCL objective}
CVCL aligns utterances to anchor frames with a symmetric contrastive loss~\citep{Vong2024GroundedLanguage}.
Define
\begin{align}
S_{ij}^{\mathrm{glob}} &= \frac{s(u_i, v_j)}{\tau_{\mathrm{glob}}},
\label{eq:glob_sim}
\end{align}
and optimize
\begin{align}
\mathcal{L}_{\mathrm{glob}}
&=
\frac{1}{2B}\sum_{i=1}^{B}
\Big(
\mathrm{CE}(S^{\mathrm{glob}}_{i,:}, i)
+
\mathrm{CE}((S^{\mathrm{glob}})^{\top}_{i,:}, i)
\Big),
\label{eq:glob_loss}
\end{align}
with temperature $\tau_{\mathrm{glob}}>0$.

\subsection{Region candidates and object embeddings}
For each frame $x_{i,m}$ we use a set of binary masks
$\mathcal{M}_{i,m}=\{M_{i,m,r}\}_{r=1}^{R_{i,m}}$
and downsample them to feature-map resolution, providing
$\tilde{M}_{i,m,r}\in[0,1]^{H\times W}$.
Let $F_{i,m}^{p,q}\triangleq F_{i,m}(:,p,q)\in\mathbb{R}^{C}$ denote the feature vector at location $(p,q)$.

\paragraph{Masked average pooling.}
We compute a region descriptor by masked averaging:
\begin{align}
f_{i,m,r}
&=
\frac{\sum_{p,q} \tilde{M}_{i,m,r}[p,q]\;F_{i,m}^{p,q}}
{\sum_{p,q}\tilde{M}_{i,m,r}[p,q] + \varepsilon}
\in\mathbb{R}^{C},
\label{eq:masked_pool}
\end{align}
with $\varepsilon>0$ for numerical stability.
A projection head $\Pi:\mathbb{R}^{C}\rightarrow\mathbb{R}^{d}$ maps region descriptors to the shared space:
\begin{align}
o_{i,m,r}=
\frac{\Pi(f_{i,m,r})}{\|\Pi(f_{i,m,r})\|_2}
\in\mathbb{S}^{d-1}.
\label{eq:obj_embed}
\end{align}

\paragraph{Patch candidates for missing or degenerate masks.}
If a mask becomes empty after downsampling, we replace it with the feature at its centroid location (in feature-map coordinates).
If a frame has no surviving masks, we select $K_{\mathrm{patch}}$ salient spatial locations from $F_{i,m}$ and treat their projected features as patch candidates.
This guarantees every frame provides a valid candidate set.

\subsection{Object files via short-window tracking}
We merge per-frame candidates into short object files using greedy similarity tracking.
Within each example $i$, candidates are assigned to the most similar existing track if cosine similarity exceeds a threshold; otherwise a new track is created.
Each track embedding is the $\ell_2$-normalized mean of its assigned candidate embeddings.
We keep at most $R_{\max}$ tracks and denote track embeddings by
$\tilde{\mathcal{O}}_i=\{\tilde{o}_{i,1},\dots,\tilde{o}_{i,R_i}\}$.

\paragraph{Null track and padding.}
We append a learnable null embedding $\tilde{o}_{\varnothing}\in\mathbb{S}^{d-1}$ to represent ``no grounded referent in the window''.
For batching, each bag is padded to size $R_{\max}+1$ with a validity mask $m_{i,r}\in\{0,1\}$ over padded entries (the null entry is always valid).

\subsection{Prototype memory}
We maintain a prototype codebook $\mathcal{P}=\{p_k\}_{k=1}^{K}\subset\mathbb{S}^{d-1}$.
Given $x\in\mathbb{S}^{d-1}$, the soft prototype assignment is:
\begin{align}
q(x)_k
=
\mathrm{softmax}_k\!\left(\frac{s(x,p_k)}{\tau_{\mathrm{proto}}}\right),
\qquad k\in\{1,\dots,K\},
\label{eq:proto_assign}
\end{align}
with temperature $\tau_{\mathrm{proto}}>0$.
We use $q_i^{\mathrm{txt}}=q(u_i)$ and $q_{i,r}^{\mathrm{trk}}=q(\tilde{o}_{i,r})$.
Prototypes are updated online by EMA using Sinkhorn-normalized assignments as in SwAV~\citep{Caron2020SwAV}.

\subsection{Prototype-space MIL alignment}
We align each utterance to the bag of object files via MIL~\citep{Maron1998MIL}.
Define instance similarity
\begin{align}
a_{ijr} = (q_i^{\mathrm{txt}})^\top q_{j,r}^{\mathrm{trk}},
\end{align}
and aggregate over tracks with masked log-sum-exp pooling:
\begin{align}
S_{ij}^{\mathrm{mil}}
=
\tau_{\mathrm{MIL}}\,
\log\!\sum_{r=1}^{R_{\max}+1}
m_{j,r}\,
\exp\!\left(\frac{a_{ijr}}{\tau_{\mathrm{MIL}}}\right),
\label{eq:mil_sim}
\end{align}
where $\tau_{\mathrm{MIL}}>0$ controls pooling hardness.
We apply a symmetric in-batch contrastive loss:
\begin{align}
\mathcal{L}_{\mathrm{mil}}
=
\frac{1}{2B}\sum_{i=1}^{B}
\Big(
\mathrm{CE}(S^{\mathrm{mil}}_{i,:}, i)
+
\mathrm{CE}((S^{\mathrm{mil}})^{\top}_{i,:}, i)
\Big).
\label{eq:mil_loss}
\end{align}

\subsection{Regularizers}

\paragraph{Track coherence.}
Object files are intended to represent the same underlying entity across frames.
For each example $i$, let $\mathcal{T}_i$ be its set of non-null tracks.
For each track $r\in\mathcal{T}_i$, let $\mathcal{A}_{i,r}\subseteq\{0,\dots,M-1\}$ be the set of frames in which the track has an assigned per-frame candidate (from the tracking step), and let $q_{i,m,r}$ be the prototype assignment of that per-frame candidate.
We regularize the per-frame assignments to match the track assignment using KL divergence with a stop-gradient teacher:
\begin{align}
\ell^{\mathrm{coh}}_{i,r}
&=
\frac{1}{|\mathcal{A}_{i,r}|}
\sum_{m\in\mathcal{A}_{i,r}}
\mathrm{KL}\!\left(\mathrm{sg}[q_{i,r}^{\mathrm{trk}}] \,\|\, q_{i,m,r}\right),
\nonumber\\
\ell^{\mathrm{coh}}_{i}
&=
\frac{1}{|\mathcal{T}_i|}
\sum_{r\in\mathcal{T}_i}
\ell^{\mathrm{coh}}_{i,r},
\nonumber\\
\mathcal{L}_{\mathrm{coh}}
&=
\frac{1}{B}\sum_{i=1}^{B}\ell^{\mathrm{coh}}_{i},
\label{eq:coherence}
\end{align}
where $\mathrm{sg}[\cdot]$ denotes stop-gradient.

\begin{table*}[t]
\centering
\footnotesize
\setlength{\tabcolsep}{1.5pt}
\renewcommand{\arraystretch}{1.15}
\begin{tabular}{l>{\centering\arraybackslash}p{0.7cm}>{\centering\arraybackslash}p{1cm}>{\centering\arraybackslash}p{0.7cm}>{\centering\arraybackslash}p{0.7cm}>{\centering\arraybackslash}p{1cm}>{\centering\arraybackslash}p{1.3cm}>{\centering\arraybackslash}p{1cm}>{\centering\arraybackslash}p{1cm}>{\centering\arraybackslash}p{1cm}>{\centering\arraybackslash}p{1cm}>{\centering\arraybackslash}p{1cm}>{\centering\arraybackslash}p{1cm}>{\centering\arraybackslash}p{1cm}>{\centering\arraybackslash}p{1cm}>{\centering\arraybackslash}p{1cm}}
\toprule
\textbf{Method} &
{\textbf{Ball}} &
{\textbf{Basket}} &
{\textbf{Car}} &
{\textbf{Cat}} &
{\textbf{Chair}} &
{\textbf{Computer}} &
{\textbf{Crib}} &
{\textbf{Door}} &
{\textbf{Foot}} &
{\textbf{Hand}} &
{\textbf{Paper}} &
{\textbf{Puzzle}} &
{\textbf{Stairs}} &
{\textbf{Table}} &
{\textbf{Window}} \\
\midrule
CVCL &
\textbf{86} & 29 & 94 & \textbf{58} & 66 &
94 & 94 & 74 & 26 & 23 &
78 & \textbf{91} & \textbf{87} & \textbf{80} & 73 \\
\textsc{BabyMind} &
75 & \textbf{42} & \textbf{95} & 56 & \textbf{70} &
94 & \textbf{98} & \textbf{78} & \textbf{39} & \textbf{31} &
\textbf{86} & 87 & 85 & 76 & \textbf{80} \\
\bottomrule
\end{tabular}
\caption{\textbf{Per-category accuracy on Labeled-S 15.} 4-way forced-choice accuracy (\%).}
\label{tab:labeleds15_perclass}
\end{table*}

\paragraph{Global-object agreement.}
CVCL-style evaluation uses the global embedding $v_i$.
To transfer object-file structure into this representation, we select the best non-null track
\begin{align}
r_i^\star
=
\arg\max_{r\in\mathcal{T}_i}
(q_i^{\mathrm{txt}})^\top q_{i,r}^{\mathrm{trk}},
\label{eq:best_track}
\end{align}
reconstruct a continuous embedding from its prototype mixture,
\begin{align}
\hat{o}_i
=
\frac{\sum_{k=1}^{K} q(\tilde{o}_{i,r_i^\star})_k\; p_k}
{\left\|\sum_{k=1}^{K} q(\tilde{o}_{i,r_i^\star})_k\; p_k\right\|_2},
\label{eq:proto_recall}
\end{align}
and minimize cosine distance:
\begin{align}
\mathcal{L}_{\mathrm{go}}
=
\frac{1}{B}\sum_{i=1}^{B}\big(1 - s(v_i,\hat{o}_i)\big).
\label{eq:go_loss}
\end{align}

\subsection{Training objective}
We optimize $(\theta,\phi)$, the projection head $\Pi$, and the null embedding.
The prototype memory is updated online as described above.
The full objective is:
\begin{align}
\mathcal{L}
=
\lambda_{\mathrm{glob}}\,\mathcal{L}_{\mathrm{glob}}
+
\lambda_{\mathrm{MIL}}\,\mathcal{L}_{\mathrm{mil}}
+
\lambda_{\mathrm{coh}}\,\mathcal{L}_{\mathrm{coh}}
+
\lambda_{\mathrm{go}}\,\mathcal{L}_{\mathrm{go}},
\label{eq:full_objective}
\end{align}
with nonnegative weights $\lambda_{\cdot}$.
Unless otherwise stated, evaluation uses the global embedding $v$ (the CVCL interface).

\section{Experiments and Analyses}
\label{sec:experiments}

We evaluate \textsc{BabyMind} in the Child-View Contrastive Learning (CVCL) setting~\citep{Vong2024GroundedLanguage}: self-supervised learning from egocentric developmental video paired with caregiver speech (SAYCam-S~\citep{Sullivan2021SAYCam}), followed by CVCL-style 4-way forced-choice tests of vision-language alignment.
Our experiments ask:
(i) does object-file supervision improve SAYCam grounding?
(ii) does it generalize under CVCL's in-vocabulary (IV) OOD protocol? and
(iii) what diagnostic structure emerges in the learned object/prototype representations?

\paragraph{Evaluation (CVCL forced-choice).}
Each trial contains a target word and four candidate images (one target, three foils). The model selects the image whose embedding has the highest cosine similarity to the text embedding; we report average accuracy (\%), with chance at $25\%$.

\subsection{Training and evaluation setup}
\label{sec:train_setup}
\paragraph{Implementation details.}
All models are implemented in PyTorch using PyTorch Lightning and trained with DistributedDataParallel on 4 AMD MI210 GPUs (batch size 8 per GPU; global batch size 32).
For both the global CVCL loss and the prototype-space MIL loss, we gather all embeddings across GPUs (along with the MIL validity masks) to utilize a shared in-batch negative set.
Precomputed AMG masks are loaded from a cached prepacked format, and we apply the same random geometric augmentations jointly to frames and masks via an image-mask transform (non-geometric appearance augmentations are applied to images only).
We reproduce the CVCL baseline within the same codebase and distributed setup for a controlled comparison.
Unless otherwise stated, for all the experiments (including CVCL reproduced results), we use AdamW with a learning rate $4\times 10^{-4}$ and weight decay $0.1$, together with global temperature $\tau_{glob}=0.07$ as in CVCL, and report seed-0 results from the checkpoint with the best validation loss.

\paragraph{Utterance-frame pairing.}
We follow CVCL's preprocessing and pairing procedure~\citep{Vong2024GroundedLanguage}:
frames are sampled at 5 FPS, and each utterance is paired with all frames until the next utterance (capped at 32 frames).
For the global CVCL objective, we sample a single \emph{anchor} frame uniformly at random from this utterance-aligned set,
exactly as in CVCL.
\textsc{BabyMind} uses the same anchor frame for $\mathcal{L}_{\mathrm{glob}}$ and additionally samples $M{-}1$ extra frames
from the same utterance-aligned set to form a $M$-frame window used only by the object-file pathway.
For tracking, we process the window frames in chronological order.
\begin{table}[t]
\centering
\small
\setlength{\tabcolsep}{6pt}
\begin{tabular}{lcc}
\toprule
\textbf{Method} & \textbf{Avg (\%)} & $\Delta$ \\
\midrule
CVCL (reprod) & 70.20 & - \\
\textsc{BabyMind} & $\mathbf{72.80}$ & $\mathbf{+2.60}$ \\
\bottomrule
\end{tabular}
\caption{\textbf{Labeled-S 15 (SAYCam-S).} 4-way forced-choice average accuracy. $\Delta$ is relative to CVCL.}
\label{tab:labeleds15_main}
\end{table}
\begin{table}[t]
\centering
\small
\setlength{\tabcolsep}{8pt}
\begin{tabular}{lc}
\toprule
\textbf{Variant} & \textbf{Avg (\%)} \\
\midrule
CVCL & 70.20 \\
\textsc{BabyMind} (full) & \textbf{72.80} \\
\midrule
w/o prototype MIL (object-centric branch) & 70.30 \\
w/o object files (no tracking) & 72.10 \\
w/o track coherence ($\lambda_{\text{coh}}{=}0$) & \underline{72.50} \\
w/o global-object agreement ($\lambda_{\text{go}}{=}0$) & 71.20 \\
\bottomrule
\end{tabular}
\caption{\textbf{Ablations on Labeled-S 15.} 4-way forced-choice average accuracy.}
\label{tab:ablations}
\end{table}
\begin{table*}[h]
\centering
\small
\setlength{\tabcolsep}{6pt}
\begin{tabular}{llccccc}
\toprule
\textbf{Dataset} & \textbf{Method} &
\textbf{\#IV} & \textbf{\#OOV} &
\textbf{OOD Avg (\%)} & \textbf{OOV Avg (\%)} & \textbf{Total Avg (\%)} \\
\midrule
\multirow{2}{*}{Konkle Object Categories}
& CVCL & 65 & 135 & \underline{34.32} & \textbf{53.11} & \underline{47.00} \\
& \textsc{BabyMind} & 65 & 135 & \textbf{35.41} & \underline{52.87} & \textbf{47.20} \\
\midrule
\multirow{2}{*}{COCO (categories)}
& CVCL & 70 & 10 & \underline{32.78} & \textbf{42.50} & \underline{34.00} \\
& \textsc{BabyMind} & 70 & 10 & \textbf{33.14} & \underline{42.36} & \textbf{34.29} \\
\bottomrule
\end{tabular}
\caption{\textbf{OOD (IV) and OOV generalization.}
IV/OOD results use the CVCL 4-way forced-choice protocol with global embeddings. \#IV/\#OOV are label counts after the vocabulary filter; Total Avg is label-count-weighted over IV and OOV.}
\label{tab:ood_oov}
\end{table*}
\subsection{Generalization under CVCL IV/OOV protocol}
\label{sec:ood_oov}
\paragraph{Object candidates.}
We precompute SAM-style automatic mask generation (AMG) proposals offline. For each frame, we load up to 24 masks, downsample them to the backbone feature-map resolution, and filter tiny masks (area $<1\%$ of the feature-map grid). If a frame has no valid masks after filtering, we construct patch candidates from the same feature map (top-$K_{\text{patch}}{=}4$ locations; patch radius 1 in feature map coordinates), ensuring that every frame yields a valid candidate set.

\paragraph{\textsc{BabyMind} hyperparameters.}
Prototype memory: $K{=}64$, $\tau_{\text{proto}}{=}0.07$, EMA decay 0.99 with Sinkhorn balancing (3 iterations, $\epsilon{=}0.05$).
Object files: greedy tracking threshold 0.55, maximum of 16 tracks per window.
Loss weights: $\lambda_{\text{glob}}{=}1.0$. $\lambda_{\text{MIL}}{=}0.10$, $\lambda_{\text{coh}}{=}0.05$, $\lambda_{\text{go}}{=}0.05$, with $\tau_{\text{MIL}}{=}0.05$.

\subsection{SAYCam grounding on Labeled-S 15}
\label{sec:labeleds15}

Our primary SAYCam evaluation is CVCL's manually filtered \textbf{Labeled-S 15} benchmark~\citep{Vong2024GroundedLanguage} (15 mutually exclusive categories; 100 forced-choice trials per category, 1500 total).
\textsc{BabyMind} improves average accuracy by $+2.60$ points over CVCL (Table~\ref{tab:labeleds15_main}).
Gains are concentrated in categories where the referent is often small, partially visible, or embedded in clutter (e.g., \texttt{Basket}, \texttt{Foot}, \texttt{Hand}, \texttt{Paper}, \texttt{Window}; Table~\ref{tab:labeleds15_perclass}), consistent with object-file MIL providing a cleaner supervisory signal than whole-frame alignment.
We observe decreases on \texttt{Ball} (and modest drops on \texttt{Puzzle}/\texttt{Table}), suggesting that when global appearance cues are already strong, imperfect masks/tracks can introduce competing gradients.

\subsection{Ablations on SAYCam-S}
\label{sec:ablations}

We ablate \textsc{BabyMind} components on Labeled-S 15 by removing each mechanism while keeping the rest fixed (Table~\ref{tab:ablations}).
Two results stand out.
First, removing prototype-space MIL (the object-centric branch) reduces performance to the CVCL baseline, showing that improvements are driven by object-file supervision rather than incidental training differences.
Second, global-object agreement is the most important auxiliary term for downstream performance: disabling it costs 1.6 points, consistent with its role in transferring object-file structure into the global embedding used by forced-choice evaluation.
Tracking into object files provides a smaller but consistent gain (0.7 points), and track coherence contributes mild regularization (0.3 points).

We evaluate beyond SAYCam-S on Konkle Object Categories and COCO categories~\citep{Lin2014COCO} using CVCL’s in-vocabulary (IV) vs.\ out-of-vocabulary (OOV) split.
IV labels use the same 4-way forced-choice test as on SAYCam-S (global embeddings), while OOV labels use a CLIP Dissect-style unit probing protocol~\citep{OikarinenWeng2022CLIPDissect}.
For Konkle, we cap at 200 images per label and use 5 repeats per image (resampling foils).
For COCO, we cap at 200 instances per label and evaluate instance crops, filtering by minimum box area $(32\times32)$ and minimum side length (16 px).
\textsc{BabyMind} yields small but consistent IV/OOD gains on both datasets with essentially unchanged OOV probing (Table~\ref{tab:ood_oov}), matching its design: object-file supervision primarily strengthens the global embedding used by forced-choice recognition.

\subsection{Qualitative and diagnostic analyses}
\label{sec:qual_analysis}

\begin{figure}[t]
    \centering
    \includegraphics[trim={5cm 5cm 5cm 5cm},width=\linewidth]{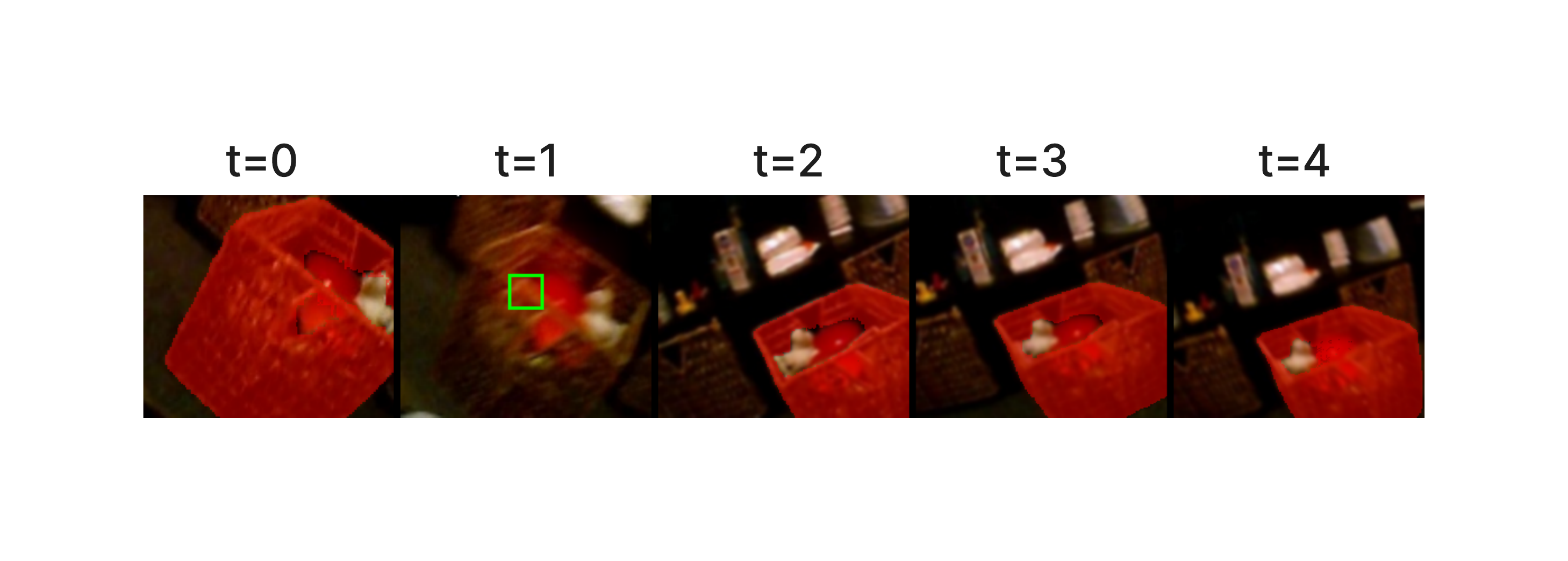}
    \caption{\textbf{Object-file robustness under mask dropouts.} A 5-frame window ($M{=}5$) for one tracked object file. Red overlays indicate selected AMG mask candidates when available. At $t{=}1$, no valid mask remains after filtering; the model falls back to a patch candidate (green box), preserving temporal coherence of the track.}
    \label{fig:track_fallback}
\end{figure}

\paragraph{Tracking through mask dropouts.}
Figure~\ref{fig:track_fallback} illustrates one tracked object file across a 5-frame window.
When AMG masks are unavailable after filtering (here at $t{=}1$), \textsc{BabyMind} falls back to a patch candidate (green box), maintaining a coherent track and keeping the object-file pathway well-defined under common egocentric failures (motion blur, occlusion, off-center framing).
\begin{figure}[t]
    \centering
    \includegraphics[width=\linewidth]{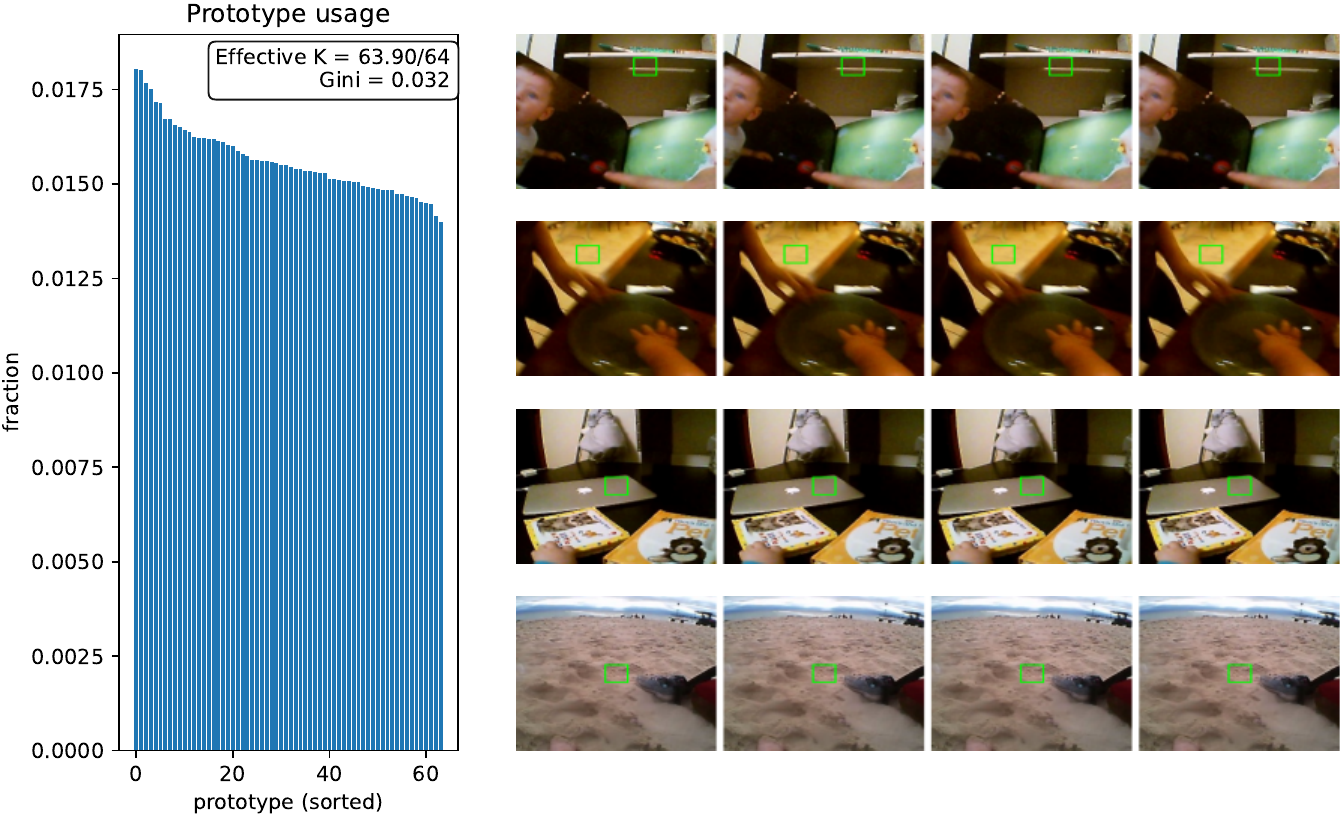}
    \caption{\textbf{Prototype diagnostics.} Left: prototype usage histogram (sorted), with effective \#prototypes and Gini coefficient. Right: for several frequently used prototypes, the top-activating tracked object file mined from training data, shown as a 4-frame strip.}
    \label{fig:proto_diag}
\end{figure}
\paragraph{Prototype memory diagnostics.}
Figure~\ref{fig:proto_diag} summarizes prototype behavior: the usage distribution (effective \#$K$ and Gini) and example top-activating tracked object files for several prototypes.
Prototype usage remains well-spread (no collapse), and the retrieved tracks show distinct recurring appearance/region patterns rather than a single dominant code.

\section{Conclusion}
We introduced \textsc{BabyMind}, an object-first inductive bias for learning grounded word meaning from child-view video paired with sparse caregiver speech. Rather than treating utterance-frame alignment as a single-frame problem, \textsc{BabyMind} uses an offline region interface, merges candidates across a short utterance-centered window into lightweight \emph{object files}, and aligns language to these latent candidates via a prototype-space multiple-instance contrastive objective. Two simple regularizers, track coherence and global-object agreement, help stabilize the object pathway and transfer its signal into the global embedding used by CVCL-style evaluation. Empirically, \textsc{BabyMind} improves SAYCam-S grounding on Labeled-S~15 and yields consistent (if modest) gains on in-vocabulary OOD benchmarks under the CVCL protocol, while leaving OOV unit-probing performance largely unchanged. Overall, the results support an ``objects before words'' perspective: even coarse, short-window perceptual organization can reduce temporal and spatial ambiguity in egocentric data and provide cleaner targets for early word grounding. Future work should reduce reliance on offline masks, extend object persistence beyond short windows, and evaluate robustness across additional children, environments, and more complex linguistic contexts.

\printbibliography

\end{document}